%% file: acl_latex.tex
\pdfoutput=1

\documentclass[11pt, table, xcdraw]{article}
\usepackage{multicol,multirow}
\usepackage{graphicx}
\usepackage{amsmath}
\usepackage[]{acl}

\usepackage{times}
\usepackage{latexsym}

\usepackage[T1]{fontenc}

\usepackage[utf8]{inputenc}
\usepackage{textcomp}
\usepackage{subfigure}
\usepackage{array}
\usepackage{soul}
\newcommand{\STAB}[1]{\begin{tabular}{@{}c@{}}#1\end{tabular}}
\usepackage{microtype}

\title{Beyond Borders: Investigating Cross-Jurisdiction Transfer in\\ Legal Case Summarization}

\author{Santosh T.Y.S.S$^{1}$,Vatsal Venkatkrishna$^{2}$, Saptarshi Ghosh$^{2}$, Matthias Grabmair$^1$\\
$^{1}$Technical University of Munich, Germany \\  
$^{2}$Indian Institute of Technology, Kharagpur, India \\
\texttt{\{santosh.tokala, matthias.grabmair\}@tum.de} \\
\texttt{vatsalvenkatkrishna@gmail.com, saptarshi@cse.iitkgp.ac.in}\\
}

\begin{document}
\maketitle
\begin{abstract}
Legal professionals face the challenge of managing an overwhelming volume of lengthy judgments, making automated legal case summarization crucial. However, prior approaches mainly focused on training and evaluating these models within the same jurisdiction. In this study, we explore the cross-jurisdictional generalizability of legal case summarization models. 
Specifically, we explore how to effectively summarize legal cases of a target jurisdiction where reference summaries are \textit{not} available. 
In particular, we investigate whether supplementing models with unlabeled target jurisdiction corpus and extractive silver summaries obtained from unsupervised algorithms on target data enhances transfer performance.
Our comprehensive study on three datasets from different jurisdictions highlights the role of pre-training in improving transfer performance. We shed light on the pivotal influence of jurisdictional similarity in selecting optimal source datasets for effective transfer. Furthermore, our findings underscore that incorporating unlabeled target data yields improvements in general pre-trained models, with additional gains when silver summaries are introduced. This augmentation is especially valuable when dealing with extractive datasets and scenarios featuring limited alignment between source and target jurisdictions. Our study provides key insights for developing adaptable legal case summarization systems, transcending jurisdictional boundaries. 
\end{abstract}

\section{Introdution}
\input{text/introduction}

\section{Related Work}

\input{text/related}

\section{Datasets}
\label{data-char}
\input{text/datasets}

\section{Models \& Evaluation Metrics}
\input{text/models}

\section{RQ1: Cross-Jurisdiction Generalizability}
\label{RQ1-sec}
\input{text/RQ1}

\section{RQ2: Leveraging Unlabelled Target Jurisdiction Corpus}
\input{text/RQ2}

\section{RQ3: Leveraging Silver Summaries of Target Jurisdiction}
\input{text/RQ3}

\section{Case Study}
\input{text/casestudy}

\section{Conclusion}
\input{text/conclusion}

\section*{Limitations}
\input{text/limitations}

\section*{Ethics Statement}
\input{text/ethics}

\bibliography{custom}
\bibliographystyle{acl_natbib}

\appendix
\input{text/appendix}

\end{document}

%% file: text/introduction.tex
Legal professionals, including lawyers, judges, and paralegals, are often inundated with an overwhelming amount of case judgements containing intricate textual content encompassing multiple issues and arguments, 
making it time-consuming to read and comprehend every document in detail manually. 
To address this, many legal information systems offer case summaries authored by legal experts. 
There has been a lot of academic research over the years to automatically produce the summaries of legal case judgements, which can help reduce the human cost effectively~\cite{bhattacharya2021incorporating,shukla2022legal,agarwal2022extractive}.

Traditional legal case judgement summarization methods have primarily focused on extractive summarization, where crucial sentences are selected from source documents. Initially, unsupervised approaches were prominent, tailored to capture legal discourse nuances without requiring training data. More recently, supervised extractive summarizers have emerged, relying on expert-annotated summaries for training. Recognizing the limitations of extractive methods in providing comprehensive context and coherence, researchers have shifted towards abstractive summarization, leveraging recent advances in unsupervised pre-training (e.g., \citealt{devlin2019bert,lewis2020bart}), legal domain-specific continued pre-training (e.g., \citealt{paul2022pre,chalkidis2020legal,gururangan2020don}), and summarization task-specific pre-training (e.g., \citealt{zhang2020pegasus}).

The task of summarizing legal case judgements presents distinctive challenges 
due to several factors, such as the legal domain vocabulary being slightly different in different jurisdictions (e.g., country / court), as well as the complex sentence and discourse structures influenced by various writing styles that are in use in different jurisdictions. 
The current strategy to design a legal case summarization system for a \emph{new} jurisdiction (which we call \emph{target jurisdiction}) either involves unsupervised methods or collecting expert annotated datasets to fine-tune supervised summarization models. 
Though these supervised models usually show improved performance compared to unsupervised methods, this often demands a large number of expert summaries, making it costly and less adaptable to new jurisdictions. 
This prompts us to ask \textit{how we can build an effective summarization system for a target jurisdiction without data annotation efforts}. 
In this regard, we seek to investigate whether models trained on some other jurisdiction for which training data exists (which we call as \emph{source jurisdiction}) can generate better summaries than unsupervised methods for the target jurisdiction. 
Thus, we evaluate the cross-jurisdiction generalizability of different legal case summarization systems and further propose an adversarial training-based setup to improve the cross-jurisdiction transfer performance, paving the way for the development of summarization systems that are generalizable in real-world legal scenarios.
We summarize our three primary research questions:

\noindent \textbf{RQ1:} In situations where reference summaries are unavailable to train supervised models for a specific target jurisdiction, can supervised summarization models, trained using data from a different jurisdiction (source jurisdiction) outperform unsupervised methods in generating more effective summaries? 
What criteria should guide the selection of the most suitable source jurisdiction for a specific target jurisdiction? Which models exhibit better cross-jurisdiction summarization capabilities?



\noindent \textbf{RQ2:} 
Can we leverage unlabeled judgment data (only documents, not reference summaries) from the target jurisdiction to improve transfer performance during training supervised summarization models on source jurisdiction? This approach is akin to unsupervised domain adaptation, where unlabelled target inputs are used when fine-tuning on a labelled source corpus. We employ an adversarial domain discriminator method~\cite{ganin2016domain} to learn the jurisdiction-agnostic feature representations to facilitate cross-jurisdictional transfer. 

\noindent \textbf{RQ3:} 
Can extractive \textit{silver summaries} on the target jurisdiction data, \textit{obtained using unsupervised summarization algorithms without manual human annotation}, further improve transfer performance in addition to unlabelled target jurisdiction data? These silver summaries aid the decoder to grasp semantic nuances in target jurisdiction, complementing adversarial domain discriminator.

\noindent To answer these questions, we conducted a comprehensive experimental analysis on three datasets from different jurisdictions. We find that supervised models trained on a different source jurisdiction can yield better summarization in the target jurisdiction, compared to unsupervised models.  Further, we observe that the choice of source dataset is highly influenced by jurisdictional similarity rather than the dataset size or abstractiveness (RQ1). We witness an improvement in transfer performance using domain discriminator-based adversarial training. However, this improvement is contingent on the choice of backbone model. For instance, we observe a decline in transfer performance when applied to Legal pre-trained Pegasus in contrast to improvement in general pre-trained BART model (RQ2). Finally, we notice an improvement when adding silver summaries in scenarios when the dataset is extractive and there is less similarity between source-target jurisdictions (RQ3).

%% file: text/related.tex
\noindent \textbf{Legal Case Summarization:}
Traditional works in this area 
have primarily employed extractive methods known for their faithfulness to the input case document \cite{bhattacharya2019comparative}. These encompass both unsupervised methods \cite{bhattacharya2021incorporating,polsley2016casesummarizer,farzindar2004atefeh,saravanan2006improving,Mandal2021ImprovingLC} and supervised approaches \cite{agarwal2022extractive,liu2019extracting,zhong2019automatic,xu2021toward,Mandal2021ImprovingLC}, employing diverse strategies such as knowledge engineering of domain-specific features, redundancy handling using MMR, joint multi-task learning with Rhetorical Role Labeling, leveraging argument structures, integrating document-specific catchphrases.

While extractive approaches excel in faithfully representing source document content, they face inherent challenges, including incomplete or incorrect discourse, coreference resolution issues, and a lack of context, impacting the readability of generated summaries \cite{zhang2022extractive}. Hence, there is a growing interest in exploring abstractive summarization methods for legal case summarization \cite{shukla2022legal,ray2020summarisation,schraagen2022abstractive}. \citealt{shukla2022legal} applied transformer-based pre-trained abstractive models like BART, Legal-LED, and Legal-Pegasus to legal case summarization. To address long legal documents, \citealt{moro2022semantic} proposed chunking input documents into semantically coherent segments, enabling these pre-trained transformer-based models to summarize lengthy documents without truncation. Recent efforts also aim to enhance factual accuracy in generated summaries by employing entailment modules to select faithful candidates \cite{feijo2023improving}. Additionally, there are efforts to understand the argumentative structure in the legal documents to improve the performance of legal case summarization models \cite{xu2022multi,elaraby2022arglegalsumm,elaraby2023towards}. In contrast to the approaches mentioned, which train and evaluate summarization models within specific jurisdictions, our work focuses on evaluating the cross-jurisdiction generalizability of legal case summarization systems.


\noindent \textbf{Cross Domain/Dataset Generalization:}
This assessment in NLP has been undertaken across various tasks, including constituent parsing \cite{fried2019cross}, reading comprehension \cite{talmor2019multiqa}, named entity recognition \cite{fu2020rethinking}, summarization \cite{chen2020cdevalsumm}, sentiment analysis \cite{yu2016learning} 
and relation extraction \cite{bassignana2022crossre}, among others. Recently, Thakur et al. \cite{thakur2021beir} created a robust benchmark called BEIR (Benchmarking IR), 
for  evaluation of retrieval model generalization capabilities across various domains.  

In the legal domain, Kumar et al. \cite{kumar2022towards} explored the cross-domain transferability of text generation models across four legal domains, including congressional bills, treaties, legal contracts, and privacy policies. However, the tasks they investigated within each domain exhibited significant variations in terms of generated text, such as titles in the privacy policy domain and actual summaries in congressional bills. 
In contrast, our work specifically focuses solely on summarizing legal case judgments. We consider each jurisdiction as a distinct domain characterized by its unique vocabulary and discourse writing style. Furthermore, we explore strategies to enhance cross-jurisdictional transfer performance through adversarial learning-based domain adaptation.

\vspace{1mm}
\noindent \textbf{Domain Adaptation:} Domain adaptation mitigates covariate shifts between source and target data distributions \cite{ruder2019neural}. Unsupervised domain adaptation (UDA) leverages labeled source data and unlabeled target data, extensively studied in NLP and computer vision \cite{daume2009frustratingly,sun2016return,wang2019adversarial,shah2018adversarial,ganin2016domain,shen2018wasserstein}. Recently, DA has extended to information retrieval \cite{xin2022zero,wang2022gpl,yu2022coco}, aided by benchmarks like BEIR. Domain adaptation often employs domain discriminator-based adversarial training \cite{ganin2016domain} to align source and target distributions within an encoder, reducing distribution disparities. In the legal domain, this technique has been applied to case outcome classification, treating legal articles as separate domains to assess transfer performance across violation predictions for different articles \cite{tyss2023zero}.


We employ DA to create jurisdiction-invariant representations in legal case summarization. However, relying solely on a jurisdiction-agnostic encoder representation, as commonly used in classification tasks, is insufficient as in the context of summarization, the decoder must also capture semantic nuances to generate domain-specific summaries. Therefore, we explore the use of silver summaries derived from unsupervised methods for target domain data.

%% file: text/datasets.tex
There are very few publicly available English legal case summary corpora available. We use the following three datasets from prior works.

\noindent \textbf{(i) UK-Abstractive dataset (abbreviated as `UK-Abs')} \cite{shukla2022legal}: This dataset includes 793 case documents and their summaries from the UK Supreme Court, spanning from 2009 onwards. The abstractive summaries are sourced from the official press-released version available on the court's website. We use the provided split, with 693 documents for training and 100 for testing. 

\noindent \textbf{(ii) Indian-Abstractive dataset (IN-Abs)} \cite{shukla2022legal}: This dataset comprises case-summary pairs from Indian Supreme Court judgements and is obtained from the website of the Legal Information Institute of India. These abstractive summaries, also known as headnotes, are available for 7130 cases, out of which 7030 are used for training and the remaining 100 for testing.

\noindent \textbf{(iii) BVA-Extractive (BVA-Ext)} \cite{zhong2019automatic}: It comprises case-summary pairs focusing on single-issue Post-Traumatic Stress Disorder decisions from the US Board of Veterans' Appeals (BVA). It encompasses a total of 112 decisions, each accompanied by an expert-annotated gold-standard extractive summary. Within this dataset, 92 cases within the training set are associated with one annotated extractive summary each. Additionally, the test set comprises 20 cases, for which four distinct extractive summaries are provided by multiple annotators. In all our experiments, for each document in the test set, we report the averaged metric score across these four reference summaries.

\vspace{2mm}
\noindent \textbf{Dataset characteristics:}
To gain deeper insights into model performance, we measure the following dataset characteristics. 
\textbf{Compression Ratio}~\cite{grusky2018newsroom} indicates the word ratio between the summary and the case document. 
\noindent \textbf{Coverage}~\cite{grusky2018newsroom} indicates the percentage of words in the summary that are part of an extractive fragment with the input, and quantifies the extent to which a summary is derivative of an input text. 
\textbf{Density}~\cite{grusky2018newsroom} quantifies how well the word sequence of a summary can be described as a series of extractions. It is derived from the average length of the extractive fragment to which each word in the summary belongs. 
\textbf{Copy Length}~\cite{chen2020cdevalsumm} denotes the average length of segments in summary copied from the source document. 
\textbf{Repetition}~\cite{see2017get} indicates the proportion of n-grams repeated in the summary itself. 
\textbf{Novelty}~\cite{see2017get} denotes the proportion of n-grams present in the summary that are \textit{not} in the source. High values of coverage, density, and copy length suggest a more extractive dataset, while novelty reflects the level of abstractiveness in the reference summaries. The lower the compression ratio, the more precise capture of critical aspects from the article text is required, which presents a greater challenge in summarization.

\input{text/tab-dataset-char}

Table~\ref{dataset-abs} reports the characteristics of the three datasets.
Barring the completely extractive BVA-Ext, UK-Abs tend to have slightly higher coverage than IN-Abs. By contrast, IN-Abs has
more copy length and density, suggesting a higher presence of verbatim copy segments when compared to UK-Abs. Conversely, UK-Abs is characterized by a higher 3-gram novelty score, which, coupled with its low compression ratio, indicates its challenging nature. Additionally, IN-Abs tends to include more 2-gram repeated phrases within the summaries. 

\vspace{2mm}
\noindent \textbf{Similarity between the datasets from different jurisdictions:}
We seek to quantify the similarity between jurisdictions on both lexical and semantic levels. On the lexical level, we calculate the \textbf{vocabulary overlap} (\%) between the jurisdictions from both the documents and summaries. 
The vocabulary for each domain is created by considering the top 1K most frequent words (excluding stopwords)~\cite{gururangan2020don} from the documents and the summaries. 
At the semantic level, we employ a pre-trained GPT-2 model~\cite{radford2019language} to compute \textbf{perplexity scores} using a strided sliding window approach with a stride of 512. We gauge the similarity between jurisdictions based on the order of perplexity scores among the jurisdictions, which provides insight into their semantic resemblance.

From Table~\ref{dataset-perp} and Table~\ref{dataset-vocab}, it becomes evident that IN-Abs holds more resemblance to UK-Abs compared to BVA-Ext. This observation can be attributed to two key factors. Firstly, India's historical ties to the UK have led to a substantial influence on its legal system, stemming from British colonial rule. Secondly, both UK-Abs and IN-Abs datasets originate from Supreme Courts dealing with a broad spectrum of legal matters due to their comprehensive jurisdiction. In contrast, the BVA corpus, focusing specifically on single-issue PTSD decisions, might exhibit more narrowly focused language. With respect to BVA-Ext, UK-Abs tend to be more similar than IN-Abs on both similarities.

%% file: text/tab-dataset-char.tex
\begin{table*}[tbp]
\begin{minipage}{0.45\linewidth}
  \centering
  \caption{Statistics of the case summarization datasets. Sum.:Summary; Doc.:Document}
  \scalebox{0.85}{
\begin{tabular}{|c|c|c|c|}
\hline
                      & UK-Abs & IN-Abs & BVA-Ext    \\ \hline
Train size            & 693    & 7030   & 92     \\ \hline
Test Size             & 100    & 100    & 20     \\ \hline
Compression Ratio     & 0.129  & 0.224  & 0.152  \\ \hline
Avg. \#Tokens in Sum. & 1268   & 1067   & 259    \\ \hline
Avg. \#Tokens in Doc. & 14599  & 5408   & 2548   \\ \hline
Copy Length           & 2.958  & 3.725  & 39.86 \\ \hline
Coverage              & 0.968  & 0.943  & 1.00   \\ \hline
Density               & 8.129  & 10.314 & 53.91 \\ \hline
Novelty (3-gram)             & 0.44   & 0.375  & 0.00  \\ \hline
Repetition (2-gram)          & 0.019  & 0.021  & 0.013  \\ \hline
\end{tabular}}
  \label{dataset-abs}
\end{minipage}%
\hfill
\begin{minipage}{0.45\linewidth}
  \centering
  \caption{Perplexity score of the datasets using GPT2}
  \scalebox{0.85}{
\begin{tabular}{|c|c|}
\hline
Dataset & Perplexity \\ \hline
UK-Abs  & 16.91      \\ \hline
IN-Abs  & 17.81      \\ \hline
BVA-Ext & 14.74      \\ \hline
\end{tabular}}
  \label{dataset-perp}
  \vspace{1em} 
  \caption{Vocabulary overlap (\%) between datasets.}
  \scalebox{0.85}{
\begin{tabular}{|c|c|c|c|}
\hline
\textbf{}        & \textbf{UK-Abs}                                     & \textbf{IN-Abs}                                     & \textbf{BVA-Ext}                                    \\ \hline
\textbf{UK-Abs}  & \cellcolor[HTML]{57BB8A}{\color[HTML]{333333} 100}  & \cellcolor[HTML]{81C897}{\color[HTML]{333333} 49.7} & \cellcolor[HTML]{CBE8D3}{\color[HTML]{333333} 30.7} \\ \hline
\textbf{IN-Abs}  & \cellcolor[HTML]{81C897}{\color[HTML]{333333} 49.7} & \cellcolor[HTML]{57BB8A}{\color[HTML]{333333} 100}  & \cellcolor[HTML]{E8F5EC}{\color[HTML]{333333} 25.9} \\ \hline
\textbf{BVA-Ext} & \cellcolor[HTML]{CBE8D3}{\color[HTML]{333333} 30.7} & \cellcolor[HTML]{E8F5EC}{\color[HTML]{333333} 25.9} & \cellcolor[HTML]{57BB8A}{\color[HTML]{333333} 100}  \\ \hline
\end{tabular}}
  \label{dataset-vocab}
\end{minipage}
\end{table*}

%% file: text/models.tex
We use the following representative models from different legal case summarization methods in \citealt{shukla2022legal} for our study. 
We briefly describe each method and refer the reader to the aforementioned work for implementation details.
\vspace{1mm}

\noindent \textbf{Unsupervised Extractive Methods:} (i) Domain-agnostic methods which employ graph-based salient sentence identification, namely LexRank \cite{erkan2004lexrank} and Reduction \cite{mihalcea2004textrank}, matrix factorization-based LSA \cite{steinberger2004using} and TF-IDF based Luhn \cite{luhn1958automatic}. 
(ii) Legal Domain-specific methods include LetSum \cite{farzindar2004atefeh}, CaseSummarizer \cite{polsley2016casesummarizer}, which ranks sentences based on their TF-IDF weights coupled with legal-specific features, Maximal Marginal Relevance (MMR) \cite{zhong2019automatic} uses iterative selection of predictive sentences and finally uses MMR to select the final summary sentences.
\vspace{1mm}

\noindent \textbf{Supervised Extractive Methods:} SummaRuNNer \cite{nallapati2017summarunner} uses binary classification for sentence selection to determine whether it should be part of the summary. 
To build the training data for the classifier from a source document $d$ and its abstractive reference summary $s$, each sentence in $s$ is matched with the top sentences from $d$, chosen based on the highest average ROUGE-1, 2 and L scores. The extractive pseudo-reference summary is created by combining all selected sentences and are then used for supervised training of the model.

\vspace{1mm}

\noindent \textbf{Abstractive Methods:} We use BART~\cite{lewis2020bart} pre-trained on general English corpora, and the
`Legal Pegasus' model\footnote{\url{https://huggingface.co/nsi319/legal-pegasus}} 
that is further trained on documents from litigations of U.S. Securities and Exchange Commission (SEC) after its pre-training phase. 
We assess their performance with and without fine-tuning. 
To accommodate the 1024-token input limit, we chunk the input document, generate summaries for each chunk, and merge them, similar to what was done in~\citealt{shukla2022legal}. 
To create reference summaries for each chunk for fine-tuning, we map each summary sentence to its most similar counterpart in the document using cosine similarity between mean token-level Sentence-BERT~\cite{reimers2019sentence} embeddings and concatenating them as references for each document chunk. 
We also use the Legal-Longformer Encoder-Decoder (Legal-LED) model\footnote{ \url{https://huggingface.co/nsi319/legal-led-base-16384}} which is trained on SEC legal corpus and is tailored specifically for long documents with 16,384 tokens through sparse attention mechanism ~\cite{beltagy2020longformer}.

\noindent \textbf{Hybrid Extractive-Abstractive Methods:} We also use both vanilla and fine-tuned models of BERT-BART~\cite{bajaj2021long}, wherein the document length is reduced by selecting salient sentences using a BERT-based extractive summarization model followed by a BART model to generate the final summary. 

\vspace{1mm}
\noindent \textbf{Evaluation metrics:} For evaluating the quality of generated and extracted summaries, we employ ROUGE-L F-score (R-L) \cite{lin2004looking} 
and BERTScore (BS) 
\cite{zhang2019bertscore} and finally report the average scores across all the test instances of the target jurisdiction. 

\input{text/tab-RQ1v4}

%% file: text/tab-RQ1v4.tex
\begin{table*}[!ht]
\centering
\caption{ROUGE-L F scores (R-L) and BERT Score (BS) of all methods across three datasets. Each test dataset is represented by a column, while training dataset used for supervision is indicated in Train column, if applicable. Best results in each group and overall are bolded in black and magenta respectively.}
\scalebox{0.95}{
\begin{tabular}{|c|c|c|c|c|}
\hline
{\textbf{}}                      & {\textbf{Test} \textrightarrow}      & {\textbf{UK-Abs}}      & {\textbf{IN-Abs}}      & \textbf{BVA-Ext}         \\ \hline
\hspace{0.7cm}{\textbf{Model}} \hspace{0.7cm}                & \hspace{0.5cm}{\textbf{Train \textdownarrow}}\hspace{0.5cm} & \hspace{0.55cm}{\textbf{R-L/BS}}\hspace{0.55cm} & \hspace{0.55cm}{\textbf{R-L/BS}}\hspace{0.55cm} & \hspace{0.55cm}\textbf{R-L/BS}\hspace{0.55cm} \\ \hline
\multicolumn{5}{|c|}{Unsupervised, Extractive Models}                                                                                                                      \\ \hline
{LexRank}                        & {-}           & {0.427/\textbf{0.836}}     & {0.439/\textbf{0.850}}     & 0.356/0.862      \\ \hline
{LSA}                            & {-}           & {0.361/0.823}     & {0.410/0.841}     & 0.308/0.854      \\ \hline
{Luhn}                           & {-}           & {0.435/0.828}     & {0.414/0.843}     & 0.353/0.860      \\ \hline
{Reduction}                      & {-}           & {0.424/0.829}     & {0.423/0.847}     & 0.364/\textcolor{magenta} {\textbf{0.865}}     \\ \hline
{CaseSummarizer}                 & {-}           & {0.443/0.835}     & {0.476/0.836}     & \textbf{0.386}/0.829      \\ \hline
{LetSum}                         & {-}           & {0.361/0.772}     & {0.362/0.797}     & 0.298/0.807      \\ \hline
{MMR}                            & {-}           & {\textbf{0.467}/0.834}     & {\textbf{0.485}/0.847}     & 0.384/0.825      \\ \hline 
\multicolumn{5}{|c|}{Supervised,  Extractive Models}                                                                                                                        \\ \hline
{\multirow{3}{*}{SummaRuNNer}}   & {UK-Abs}         & {-}     & {\textbf{0.410}/\textbf{0.832}}     & \textbf{0.405}/\textbf{0.861}      \\ \cline{2-5} 
{}                               & {IN-Abs}         & {\textbf{0.430}/\textbf{0.834}}     & {-}     & 0.327/0.846      \\ \cline{2-5} 
{}                               & {BVA-Ext}        & {0.378/0.782}      & {0.374/0.839}     & -     \\ \hline
\multicolumn{5}{|c|}{Abstractive \& Hybrid Models without fine-tuning}   \\ \hline
{BART}        & {-}        & {\textbf{0.517}/0.833}     & {0.486/0.839}     & 0.324/0.841      \\  \hline 
{Legal-Pegasus} & {-}        & {\textbf{0.517}/0.831}     & {\textbf{0.522/0.841}}     & \textbf{0.386}/0.830      \\  \hline 
{Legal-LED}   & {-}        & {0.240/0.821}     & {0.292/0.812}     & 0.232/0.832      \\ \hline 
{BERT-BART}     & {-}        & {0.501/\textbf{0.836}}     & {0.482/0.838}     & 0.329/\textbf{0.844}      \\ \hline 
\multicolumn{5}{|c|}{Abstractive \& Hybrid Models with fine-tuning}                                                                                                                                   \\ \hline
{\multirow{3}{*}{BART}}          & {UK-Abs}         & {-}     & {\textbf{0.522}/\textbf{0.845}}     & \textbf{0.341}/\textbf{0.842}      \\ \cline{2-5} 
{}                               & {IN-Abs}         & {\textbf{0.532}/\textbf{0.837}}     & {-}     & 0.306/0.834      \\ \cline{2-5} 
{}                               & {BVA-Ext}        & {0.519/0.835}     & {0.503/0.840}     & -     \\ \hline
{\multirow{3}{*}{Legal-Pegasus}} & {UK-Abs}         & {-}     & 
\textcolor{magenta} {\textbf{0.532}}/ 
\textcolor{magenta} {\textbf{0.847}} & 
\textcolor{magenta} {\textbf{0.451}}/\textbf{0.861}      \\ \cline{2-5} 
{}                               & {IN-Abs}         & 
\textcolor{magenta} {\textbf{0.533}}/ 
\textcolor{magenta} {\textbf{0.838}} & {-}     & 0.434/0.854      \\ \cline{2-5} 
{}                               & {BVA-Ext}        & {0.522/0.834}     & {0.522/0.842}     & -      \\ \hline
{\multirow{3}{*}{Legal-LED}}                                & {UK-Abs}         & {-}     & {\textbf{0.418}/\textbf{0.838}}     & 0.321/\textbf{0.834}      \\ \cline{2-5} 
{}                               & {IN-Abs}         & {\textbf{0.479}/\textbf{0.832}}     & {-}     & \textbf{0.339}/0.831     \\ \cline{2-5} 
{}                               & {BVA-Ext}        & {0.233/0.817}     & {0.289/0.812}     & -      \\ \hline
{\multirow{3}{*}{BERT-BART}}    & {UK-Abs}         & {-}     & {\textbf{0.513/0.846}}     & \textbf{0.342/0.842}      \\ \cline{2-5} 
{}                               & {IN-Abs}         & {\textbf{0.515/0.837}}     & {-}     & 0.308/0.836      \\ \cline{2-5} 
{}                               & {BVA-Ext}        & {0.508/0.836}     & {0.501/0.839}     & -     \\ \hline
\end{tabular}}
\label{RQ1}
\end{table*}

%% file: text/RQ1.tex

Unlike previous studies~\cite{bhattacharya2019comparative,shukla2022legal} that primarily evaluated these summarization algorithms in in-domain settings, where fine-tuning and evaluation occur on the same dataset, we focus on assessing their generalization capabilities across domains/jurisdictions. 
Unsupervised models can be applied directly to summarize documents of any dataset (jurisdiction).
For supervised models, we train the model on a specific dataset (source jurisdiction) and evaluate its performance on other datasets (target jurisdictions) in a blind zero-shot manner.

\vspace{1mm}
\noindent \textbf{Results} (Table \ref{RQ1}): Among unsupervised methods (Block 1 in Tab.\ref{RQ1}), domain-specific algorithms CaseSummarizer and MMR consistently performed the best, especially in terms of Rouge-L score. 
Supervised extractive model SummaRuNNer did not consistently yield better cross-jurisdiction performance compared to unsupervised models, owing to its lack of pre-training except in case of BVA-Ext target using UK-Abs as source (Block 2 in Tab. \ref{RQ1}).

Abstractive \& Hybrid models, due to their pre-training, demonstrated better performance compared to unsupervised methods even without any fine-tuning, as reflected over UK-Abs and IN-Abs primarily due to their abstractive nature of datasets and pre-training objective (Block 3 in Tab.\ref{RQ1}). LED is semi-pretrained as it has been initialized by repeatedly copying BART without undergoing end-to-end general pre-training and then it has been fine-tuned on SEC legal corpus to create Legal-LED, making it overfit to that specific domain, as reflected in its lower zero-shot performance. 

Finally, abstractive \& hybrid models fine-tuned on datasets from source jurisdictions generally performed well compared to the unsupervised methods, emphasizing the importance of legal adaptation during fine-tuning in addition to their pre-training either generally as with BERT/BART-BERT or legal-specific as with Legal-Pegasus (Block 4 in Table~\ref{RQ1}). 
They all displayed greater generalization stability across different source datasets irrespective of the relation between the source and the target. Overall, Legal-Pegasus stood out as the most robust model across all the datasets attributed to its legal pre-training. 

\vspace{0.4em}
\noindent \textbf{Choice of Source Jurisdiction:} For both UK-Abs and IN-Abs, training on each other's data yielded better results compared to training on BVA-Ext (e.g., BART applied on UK-Abs test set, achieves 0.532/0.837 when trained on IN-Abs, compared to 0.519/0.835 when trained on BVA-Ext) due to their larger size, abstractive nature, and jurisdictional similarity between In-Abs and UK-Abs, as stated earlier in Section~\ref{data-char}. Interestingly, for BVA-Ext, UK-Abs proved to be a better source domain than IN-Abs (for most supervised methods), despite the latter's larger size and more extractive nature. This transferability is mainly attributed to the similarity between jurisdictions, as observed in Section~\ref{data-char}. This result shows that, while choosing the source jurisdiction (whose data will be used to train supervised summarization models), blindly choosing the jurisdiction with the largest dataset size is not an optimal approach. Rather, the source jurisdiction should be chosen based on its lexical and semantic similarity with the target jurisdiction. To this end, the metrics we used in Section~\ref{data-char} can be useful to quantify the similarity between jurisdictions. 

\vspace{1mm}
\noindent \textbf{Main Takeaways:} 
Our findings reveal that supervised models trained on a different source jurisdiction can yield better summarization in the target jurisdiction, compared to unsupervised models.
Specifically, pre-trained models exhibit superior cross-jurisdictional generalizability due to their pre-training. Legal-oriented training further enhances their generalization capabilities. The choice of the source dataset significantly influences the development of an effective system for a specific target jurisdiction, with jurisdictional similarity playing a critical role than dataset size and abstractiveness.

%% file: text/RQ2.tex
In this section, we explore whether we can improve the cross-domain performance of supervised abstractive summarization models using the \textit{unlabeled} judgement documents from the target jurisdiction, while training the models on source jurisidiction judgement-summary pairs. 
To this end, we adopt \textit{domain-adversarial training} from unsupervised domain adaptation literature \cite{ganin2016domain}. 
Note that, we still do \textit{not} use any reference summary from the target jurisdiction. 

In this technique, we additionally introduce an \textit{auxiliary jurisdiction classifier} that tries to discriminate the source jurisdiction embeddings from the target ones. The encoder of the abstractive model is updated not only to facilitate the decoder in producing summaries for source jurisdiction documents, but is also adversarially trained to confuse the jurisdiction classifier to learn jurisdiction-invariant feature representations. Put differently, we want our models to learn how to summarize the given input document with minimal encoding of the jurisdiction-specific information contained in the texts, facilitating model to generalize to documents from target jurisdictions.

\input{text/tab-RQ2-3v6}

In our adaptation scenario, we have access to labelled judgement-summary pairs of a source jurisdiction $\{x_s,  y_s\}$ and unlabelled judgement corpus from target jurisdiction $\{x_t\}$. Let $\theta_e$ and $\theta_d$ be the encoder $E$ and decoder $D$ parameters of a summarization model. We introduce an auxiliary jurisdiction classifier $J$ with parameters $\theta_j$, implemented as a linear layer, which takes the document representation from the encoder and performs binary classification to identify the document's jurisdiction. We train the classifier in an adversarial fashion to maximize the encoder’s ability to capture information required for the summarization task while minimizing its ability to predict the jurisdiction. 
Instead of the two-step adversarial min-max objective of GAN~\cite{goodfellow2014generative}, \citealt{ganin2016domain} proposed to jointly optimize all the components using a Gradient Reversal Layer (GRL). The GRL is inserted between the encoder and jurisdiction classifier, where it acts as the identity during the forward pass but, during the backward pass, scales the gradients flowing through by $-\lambda$, making the encoder receive the opposite gradients from the jurisdiction classifier. Here, hyperparameter $\lambda$ is the \textit{adaptation rate} controlling the influence of the jurisdiction classifier on the encoder during training. Thus, the overall objective function can be compactly represented as:
\begin{equation}
   \label{grl}
   \begin{split}
   \arg \min\limits_{\theta_{e}, \theta_{d}, \theta_{j}}    L_1(D(E(x_s)), y_s) ~ + \\ \lambda L_2(J(GRL(E(x_?))), y_j)
   \end{split}
\end{equation}
where the first term represents the actual loss term for the summarization task, computed over the labelled article-summary of source jurisdiction $\{x_s,  y_s\}$ and the second term represents the binary cross entropy loss computed for classification of jurisdiction for both source and target documents $x_?$ ($x_?$ can be $x_s$ or $x_t$) and it belongs to, between the source or target jurisdiction ($y_j$).  

\vspace{1mm}
\noindent \textbf{Experimental Setup:} 
We select BART, Legal-Pegasus, and BERT-BART for these experiments, given their robust cross-domain performance in Section \ref{RQ1-sec}. 
For each pair of source-target dataset, we apply this setup to all three abstractive models. We adopt the hyperparameters from \citealt{shukla2022legal}. The jurisdiction classifier takes the mean of token-level embeddings obtained from the encoder as the document representation. The adaptation rate is given by $\lambda = \frac{2}{1+\exp(-\gamma p)}-1$, where $p = \frac{t}{T}$ and $t$ and $T$ represent the current training steps and total steps respectively. We fine-tune $\gamma$ within [0.05, 0.1] using the validation set.

\vspace{1mm}
\noindent \textbf{Results} (Table~\ref{RQ2-3}): On comparing the models trained with the GRL (RQ2) to their non-GRL counterparts (blind zero-shot transfer, as in RQ1), we notice an improvement with addition of GRL for both BART and BERT-BART across the three test sets, indicating that the models gained more robust and domain-invariant representations through adversarial training, ultimately enhancing their transferability. 
For instance, for BART trained on IN-Abs and tested on UK-Abs, the R-L/BS scores for non-GRL variant are 0.532/0.837, and they improved to 0.536/0.838 for the GRL variant.
This improvement is greatly pronounced when we consider BVA-Ext as the test set (e.g., 0.341/0.842 to 0.460/0.868 in the case of UK-Abs as the train dataset for BART).
However, in the case of Legal-Pegasus, which already demonstrated strong cross-domain generalizability (RQ1) due to its pre-training on legal texts, we notice a decline in its performance with addition of GRL (eg., 0.533/0.836 to 0.528/0.834 in case of IN-Abs as train, UK-Abs as test). We attribute this to what we refer to as `representation erasure'. It seems that the incorporation of the GRL and jurisdiction discriminator inadvertently leads to the dilution of generalizable knowledge that the model had acquired during its pre-training phase. This erasure effect is most pronounced in BVA-Ext as test dataset (e.g., 0.451/0.861 to 0.394/0.829 when UK-Abs is used as train dataset), which aligns with Legal-Pegasus pre-training data in terms of its jurisdiction (US) specific features compared to others.

\noindent \textbf{Main Takeaway:} Adversarial learning can enhance general pre-trained models (e.g., BART, BERT-BART) making them proficient at acquiring domain-invariant representations. But this strategy can also result in decreased performance for legal pre-trained models like Legal-Pegasus, attributable to representation erasure.

%% file: text/tab-RQ2-3v6.tex
\begin{table*}[!ht]
\centering
\caption{ROUGE-L F-score (R-L) and BERT Score (BS) of three abstractive summarization models trained under different settings. Each target (test) dataset is represented by a column and and 'S' columns denote the source dataset for fine-tuning. 
An empty configuration indicates that the model is fine-tuned on the source dataset and evaluated on the target dataset (RQ1). `GRL' indicates models fine-tuned in an adversarial fashion (RQ2). `Silver' indicates models trained with silver summaries of the target dataset obtained through an unsupervised MMR algorithm (RQ3). Entries marked with $^*$ are statistically significantly higher than the baseline (the empty configuration) using 95\% confidence interval by Wilcoxon signed rank test. 
}
\scalebox{1.0}{
\begin{tabular}{|cc|c|c|c|c|c|c|}
\hline
\multicolumn{1}{|c|}{}                               & \textbf{Target   \textrightarrow}   &                         & \textbf{UK-Abs}      &                         & \textbf{IN-Abs}      &                         & \textbf{BVA-Ext}         \\ \hline
\multicolumn{2}{|c|}{\textbf{Model Config}}                                        & \hspace{0.15cm}\textbf{S} \hspace{0.15cm}                      & \hspace{0.55cm}\textbf{R-L/BS} \hspace{0.55cm}     & \hspace{0.15cm}\textbf{S}\hspace{0.15cm}                       & \hspace{0.55cm}\textbf{R-L/BS}\hspace{0.55cm}      & \hspace{0.15cm}\textbf{S} \hspace{0.15cm}                      & \hspace{0.55cm}\textbf{R-L/BS}\hspace{0.55cm}      \\ \hline
\multicolumn{1}{|c|}{\multirow{8}{*}{\hspace{0.15cm}BART \hspace{0.15cm}}}  &             & \multirow{3}{*}{\STAB{\rotatebox[origin=c]{90}{IN}}}
 & 0.532/0.837 & \multirow{3}{*}{\STAB{\rotatebox[origin=c]{90}{UK}}} & 0.522/0.847 & \multirow{3}{*}{\STAB{\rotatebox[origin=c]{90}{UK}}} & 0.341/0.842 \\ \cline{2-2} \cline{4-4} \cline{6-6} \cline{8-8} 
\multicolumn{1}{|c|}{}                               & GRL         &                         & \textbf{0.536/0.838} &                         & \textbf{0.526/0.848} &                         & 0.460/0.868 $^*$ \\ \cline{2-2} \cline{4-4} \cline{6-6} \cline{8-8} 
\multicolumn{1}{|c|}{}                               & Silver &                         & 0.529/0.838 &                         & 0.507/0.832 &                         & \textbf{0.471/0.872} $^*$ \\ \cline{2-8} 
\multicolumn{1}{|c|}{}                               &             & \multirow{3}{*}{\STAB{\rotatebox[origin=c]{90}{BVA}}}   & 0.519/0.835 &\multirow{3}{*}{\STAB{\rotatebox[origin=c]{90}{BVA}}}    & 0.503/0.840  & \multirow{3}{*}{\STAB{\rotatebox[origin=c]{90}{IN}}} & 0.306/0.834 \\ \cline{2-2} \cline{4-4} \cline{6-6} \cline{8-8} 
\multicolumn{1}{|c|}{}                               & GRL         &                         & 0.544/0.837 $^*$ &                         & 0.513/0.842 $^*$ &                         & 0.440/0.864 $^*$ \\ \cline{2-2} \cline{4-4} \cline{6-6} \cline{8-8} 
\multicolumn{1}{|c|}{}                               & Silver &                         & \textbf{0.546/0.839} $^*$ &                         & \textbf{0.515/0.842} $^*$ &                         & \textbf{0.471/0.871} $^*$ \\ \hline
\multicolumn{1}{|c|}{\multirow{8}{*}{\begin{tabular}[c]{@{}c@{}}BERT-\\ BART\end{tabular}}}     &             & \multirow{3}{*}{\STAB{\rotatebox[origin=c]{90}{IN}}} & 0.515/0.838 & \multirow{3}{*}{\STAB{\rotatebox[origin=c]{90}{UK}}} & 0.513/0.846 & \multirow{3}{*}{\STAB{\rotatebox[origin=c]{90}{UK}}} & 0.342/0.842 \\ \cline{2-2} \cline{4-4} \cline{6-6} \cline{8-8} 
\multicolumn{1}{|c|}{}                               & GRL         &                         & \textbf{0.520/0.838} $^*$ &                         & \textbf{0.522/0.848}$^*$ &                         & 0.461/0.868 $^*$ \\ \cline{2-2} \cline{4-4} \cline{6-6} \cline{8-8} 
\multicolumn{1}{|c|}{}                               & Silver &                         & 0.516/0.838 &                         & 0.508/0.838 &                         & \textbf{0.465/0.868} $^*$ \\ \cline{2-8} 
\multicolumn{1}{|c|}{}                               &             & \multirow{3}{*}{\STAB{\rotatebox[origin=c]{90}{BVA}}}  & 0.508/0.837 & \multirow{3}{*}{\STAB{\rotatebox[origin=c]{90}{BVA}}}    & 0.501/0.839 & \multirow{3}{*}{\STAB{\rotatebox[origin=c]{90}{IN}}} & 0.308/0.836 \\ \cline{2-2} \cline{4-4} \cline{6-6} \cline{8-8} 
\multicolumn{1}{|c|}{}                               & GRL         &                         & 0.522/0.838 $^*$ &                         & 0.504/0.840  &                         & 0.440/0.857 $^*$ \\ \cline{2-2} \cline{4-4} \cline{6-6} \cline{8-8} 
\multicolumn{1}{|c|}{}                               & Silver &                         & \textbf{0.527/0.838} $^*$ &                         & \textbf{0.509/0.843} $^*$ &                         & \textbf{0.465/0.869} $^*$ \\ \hline
\multicolumn{1}{|c|}{\multirow{8}{*}{\begin{tabular}[c]{@{}c@{}}Legal-\\ Pegasus\end{tabular}}}  &             & \multirow{3}{*}{\STAB{\rotatebox[origin=c]{90}{IN}}} & 0.533/0.836 & \multirow{3}{*}{\STAB{\rotatebox[origin=c]{90}{UK}}} & 0.532/0.847 & \multirow{3}{*}{\STAB{\rotatebox[origin=c]{90}{UK}}} & 0.451/0.861 \\ \cline{2-2} \cline{4-4} \cline{6-6} \cline{8-8} 
\multicolumn{1}{|c|}{}                               & GRL         &                         & 0.528/0.834 &                         & 0.524/0.838 &                         & 0.394/0.829 \\ \cline{2-2} \cline{4-4} \cline{6-6} \cline{8-8} 
\multicolumn{1}{|c|}{}                               & Silver &                         & \textbf{0.534/0.836} &                         & \textbf{0.537/0.848} $^*$ &                         & \textbf{0.469/0.868} $^*$ \\ \cline{2-8} 
\multicolumn{1}{|c|}{}                               &             & \multirow{3}{*}{\STAB{\rotatebox[origin=c]{90}{BVA}}}   & 0.522/0.834 & \multirow{3}{*}{\STAB{\rotatebox[origin=c]{90}{BVA}}}    & 0.522/0.842 & \multirow{3}{*}{\STAB{\rotatebox[origin=c]{90}{IN}}} & 0.434/0.854 \\ \cline{2-2} \cline{4-4} \cline{6-6} \cline{8-8} 
\multicolumn{1}{|c|}{}                               & GRL         &                         & 0.517/0.828 &                         & 0.521/0.842 &                         & 0.326/0.818 \\ \cline{2-2} \cline{4-4} \cline{6-6} \cline{8-8} 

\multicolumn{1}{|c|}{}                               & Silver &                         & \textbf{0.524/0.835} &                         & \textbf{0.524/0.844} &                         & \textbf{0.465/0.862}  $^*$\\ \hline
\end{tabular}}
\label{RQ2-3}
\end{table*}

%% file: text/RQ3.tex
In this section, we explore whether incorporation of silver summaries from the target jurisdiction, obtained via \textit{unsupervised} extractive summarization algorithms without the need for human annotations, can enhance effectiveness of cross-domain transfer. 
We employ the same domain-adversarial training framework, with a slight modification accounting for availability of judgment-summary pairs from both the source jurisdiction (expert-annotated) and the target jurisdiction (extractive silver summaries obtained via unsupervised algorithms). 
The first term in overall objective, as described in Eqn.~\ref{grl}, is now used for target judgement-summary pairs as well. In other words, the $x_s$ in first term of Eqn.~\ref{grl} can now be $x_?$ (i.e., either $x_s$ or $x_t$). 
This modification also strengthens the robustness of the adversarial learning approach. In the previous setup, the decoder's updates were solely influenced by source jurisdiction summaries, potentially leading to overfitting to source jurisdiction semantics and limiting its generalizability to new jurisdictions. The inclusion of silver summaries helps the decoder align with the semantics of the target jurisdiction, thereby facilitating more effective transfer performance.

\vspace{1mm}
\noindent \textbf{Experimental Setup:} Similar to RQ2, we replicate the setup using three abstractive models BART, Legal-Pegasus and BERT-BART for each dataset pair, designating one as the source jurisdiction with expert-annotated summaries and the other as the target jurisdiction with silver summaries. We obtain the silver summaries for the target jurisdiction using the MMR algorithm.\footnote{We also experimented with CaseSummarizer to obtain the silver summaries and observed that MMR showed better results than CS, as the trend in RQ1 Tab.~\ref{RQ1}.}

\vspace{1mm}
\noindent \textbf{Results} (Table~\ref{RQ2-3}): With respect to BART and BERT-BART on UK-Abs (IN-Abs) as the target dataset, augmenting models with silver summaries yield improvements using BVA-Ext as train (eg., 0.544/0.837 to 0.546/0.839 on UK-Abs test, BVA-Ext train for BART), while models trained on IN-Abs (UK-Abs) did not exhibit enhancement. This can be attributed to the semantic similarity between UK-Abs and IN-Abs, allowing effective learning without additional silver summaries from each other. However, while using BVA-Ext as train dataset, the semantics specific to UK-Abs (IN-Abs) necessitate the inclusion of their silver summaries to facilitate the learning of target semantics. 

With respect to Legal-Pegasus on both UK-Abs and IN-Abs as test set, we witness improvements in all training setups. For BVA-Ext as the test set,  incorporation of silver summaries led to greater performance improvements across all three models and both source datasets consistently (e.g., 0.460/0.868 to 0.471/0.872 on UK-Abs train set on BART). This lends to the nature of BVA-Ext dataset being extractive getting boost from addition of silver summaries  which are extractive.

\noindent \textbf{Main Takeaways:} Adding silver summaries significantly improved performance, particularly for extractive datasets like BVA-Ext and when dealing with less similar jurisdictions (e.g., using BVA-Ext for training and IN-Abs/UK-Abs for testing). This, along with our RQ2 findings, highlights the efficacy of silver summaries in mitigating representational erasure in legal pre-trained models, as seen in the improved performance of Legal-Pegasus.

%% file: text/casestudy.tex
Tables~\ref{error1} and~\ref{error2} (in the Appendix) illustrate some errors found in the summaries generated by different configurations of BART models for a specific Indian Supreme Court judgement. One notable error involves Indian jurisdiction-specific terms such as I.P.C. (Indian Penal Code). Models with blind-zero transfer (RQ1) and those with access to the source text only in GRL (RQ2) often output terms like K.P.C. or K.C. (in place of I.P.C.), demonstrating a limitation in the semantics of the decoder to understand Indian-specific jargon. 
However, upon adding silver summaries (RQ3), the decoder becomes exposed to such jargon, leading to the correct output of I.P.C. 
Another common error across all models is incomplete sentences and coherence issues. In some cases, the model abruptly shifts to a new aspect without completing the previous one, potentially resulting in misrepresentations of the input document. These observations highlight the challenges in long text summarization, and the need for future improvement.

%% file: text/conclusion.tex

Our study reveals the following practical insights on building an effective legal summarization system for a target jurisdiction without annotated data: 
(i)~Fine-tuning on non-target datasets outperforms unsupervised methods, but success depends on the similarity between source and target jurisdictions. 
(ii)~When one has access to similar source data (e.g., IN-Abs for target UK-Abs), using general pre-trained models like BART and BERT-BART, combined with adversarial training, enhances transfer. 
(iii)~When access to only non-similar source data (e.g., BVA-Ext for UK-Abs) is available, augmenting models with silver summaries from the target jurisdiction improves transfer. 
(iv)~When employing adversarial learning with legal pre-trained models like Legal-Pegasus, it is important to be mindful of representational erasure, which can be mitigated by incorporating silver summaries.


%% file: text/limitations.tex
While our research contributes valuable insights into the cross-jurisdictional generalizability of legal case summarization systems, it is crucial to acknowledge certain limitations that affect the  applicability of our findings. Our experiments are conducted on three specific legal datasets from diverse jurisdictions. The extent to which our observations hold true for a broader range of legal systems remains an open question. The legal domain is vast and varied, and different jurisdictions may exhibit unique characteristics that impact the generalizability of summarization models. The scarcity of publicly available legal summarization datasets limits the diversity and size of our training and evaluation data. Hence we are constrained with use of these three, only publicly available legal case summarization datasets in English. This constraint could impact the robustness and generalizability of our proposed methods and insights.
 
Our evaluation primarily relies on established summarization metrics such as ROUGE and BERTScore. While these metrics have been used in many prior works on legal document summarization, and are known to provide a quantitative measure of summarization quality, they may not fully capture the nuanced legal content, context, and intricacies essential for legal professionals. A potential avenue for further research could be developing additional legal domain-specific evaluation metrics. Another significant limitation of our study is the absence of direct participation or validation by legal experts in the assessment of summarization outputs, which we could not perform due to lack of access to legal experts. 

%% file: text/ethics.tex
We use publicly available datasets from prior works, which are obtained from the web. Though the case documents are not anonymized, we do not foresee any harm beyond their availability. We acknowledge the potential presence of biases within the training data, which may inadvertently be reflected in the generated summaries. To deploy these models in a production system, one must thoroughly check for such biases by comprehensively evaluating summarization performance across relevant groups (e.g., gender and race). Legal systems can even carry historical biases, and training models on biased datasets may perpetuate or exacerbate existing inequalities. As such, we urge vigilance in scrutinizing dataset biases and committing to fair and unbiased representation.

An inherent challenge in the development of automatic summarization models for legal decisions lies in potential performance variations across different partitions with in the same legal domain. For instance, in contexts like the Board of Veterans' Appeals (BVA) as discussed in \citealt{agarwal2022extractive}, cases involving rarely occurring disabilities or specialized legal and military situations may lead to suboptimal summaries due to sparsity in the training data. This variability could disproportionately impact groups that should be treated equally if their characteristics coincide with these less frequent legal configurations. Engaging domain experts to curate datasets with better representation across different types of injuries and legal phenomena can be a proactive step. This can enhance the model's understanding of uncommon or group-related legal contexts, potentially mitigating disparities in performance. 

%% file: text/appendix.tex
\input{text/segment-tabv2}

\section{Implementation Details}
We follow the hyper parameters for all of the baseline models as outlined by \citealt{shukla2022legal}. All the implementational details and code of the unsupervised baselines are provided in their github repository\footnote{https://github.com/Law-AI/summarization}. For BART, we employ a learning rate of 2e-5. We train the model end-to-end for 3 epochs with a batch size of 1. We limit maximum input and output length to 1024 and 512 respectively.  For Legal Pegasus, we use a learning rate of 5e-5 and with batch size of 1 for 2 epochs, with maximum input and output length of 512 and 256 respectively. For Legal LED, we use a learning rate of 1e-3 trained for 3 epochs with a batch size of 4, with input and output length of 16384 and 1024 respectively. All models are trained with Adam optimizer \cite{kingma2014adam}. 



\input{text/tab_errors}

\section{Segment-wise Evaluation}
Each summary in UK-Abs dataset is segmented into three segments -- `Background to the Appeal', `Judgement', and `Reasons for Judgement' --allowing segment-wise evaluation to assess the quality of summarization for each rhetorical segment. In line with \citealt{shukla2022legal}, we present the ROUGE-L recall scores for each rhetorical segment, as available in the UK-Abs dataset (stated in Sec. \ref{data-char}). These scores are computed by comparing the generated summary to the corresponding golden summary under each segment. 

Table \ref{segment} showcases the results for BART and BERT-BART models trained on IN-Abs and BVA-Ext source datasets, considering both non-GRL, GRL, and silver variants. Notably, when trained with BVA-Ext, the incorporation of silver summaries yields better scores across all segments than GRL and non-GRL variants. Conversely, for IN-Abs, the GRL variant exhibits strong performance across both models than silver summary variants. These segment-level trends also align with document-wise trends in Table \ref{RQ2-3}, where performance of silver-enhanced models dropped when trained on IN-Abs but improved when trained on BVA-Ext (UK-Abs column of Tab. \ref{RQ2-3}).

%% file: text/segment-tabv2.tex
\begin{table*}[!tb]
\caption{Segment-wise ROUGE-L Recall scores on the UK-Abs dataset. BGA, Jud., Rea. indicate ‘Background to the Appeal’, ‘Judgement’, and ‘Reasons for Judgement’ respectively.}
\centering
\begin{tabular}{|c|c|c|c|c|c|c|c|}
\hline
\multicolumn{2}{|c|}{\hspace{2cm}\textbf{Source \textrightarrow}}                                                                                & \multicolumn{3}{c|}{\textbf{IN-Abs}}                                                                  & \multicolumn{3}{c|}{\textbf{BVA-Ext}}                                                                     \\ \hline
\multicolumn{2}{|c|}{\textbf{Model Config}}                                                                         & {\hspace{0.22cm}\textbf{BGA}\hspace{0.22cm}} & \hspace{0.22cm}{\textbf{Jud.}}\hspace{0.22cm} & \hspace{0.22cm}\textbf{Rea.}\hspace{0.22cm} & \hspace{0.22cm}{\textbf{BGA}}\hspace{0.22cm} & \hspace{0.22cm}{\textbf{Jud.}}\hspace{0.22cm} & \hspace{0.22cm}\textbf{Rea.}\hspace{0.22cm} \\ \hline
{\multirow{3}{*}{BART}}      &             & {0.319}               & {0.488}              & 0.283            & {0.315}               & {0.449}              & 0.295            \\ \cline{2-8} 
{}                           & GRL         & \textbf{0.377}               & \textbf{0.528}              & \textbf{0.340}             & {0.318}               & {0.452}              & 0.301            \\ \cline{2-8} 
{}                           & Silver & {0.348}               & {0.492}              & 0.293            & \textbf{0.338}               & \textbf{0.471}              & \textbf{0.302}            \\ \hline
{\multirow{3}{*}{BART-BERT}} &             & {0.312}               & {0.474}              & 0.266            & {0.309}               & {0.445}              & 0.281            \\ \cline{2-8} 
{}                           & GRL         & \textbf{0.337}               & \textbf{0.493}              & \textbf{0.294}            & {0.313}               & {0.45}               & 0.286            \\ \cline{2-8} 
{}                           & Silver & \textbf{0.337}               & {0.479}              & 0.286            & \textbf{0.334}               & \textbf{0.47}               & \textbf{0.294}            \\ \hline
\end{tabular}
\label{segment}
\end{table*}

%% file: text/tab_errors.tex
\begin{table*}[]
\centering
\begin{tabular}{|l|l|l|l|}
\hline
Config & Train & Summary Snippet & Explanation for Errors \\ \hline
\multirow{2}{*}{\begin{tabular}[c]{@{}l@{}}Blind \\ transfer \\ (RQ1)\end{tabular}} & UK-Abs  & \begin{tabular}[c]{@{}l@{}}... under \textcolor{blue}{section 302 K.P.C.} ...... \\ and had a bath there. \textcolor{magenta}{He  The} \\ \textcolor{magenta}{murder} was committed ........... \\ mere possession of a sword so \\ stained be \textcolor{teal}{not sufficient to} \\ \textcolor{teal}{establish conclusively.} The extra \\ judicial confession  of the ......\end{tabular}&    \begin{tabular}[c]{@{}l@{}}\\  \textcolor{blue}{It mentions as K.P.C., but it has to}\\ \textcolor{blue}{be I.P.C.} \\ \\ \textcolor{magenta}{Incomplete sentence: It starts with “He”} \\ \textcolor{magenta}{and digresses to the next sentence.}\\ \\ \textcolor{teal}{Incoherent discourse: It doesn’t mention} \\ \textcolor{teal}{what can’t be established. The closest} \\ \textcolor{teal}{sentence in the source document is} \\ \textcolor{teal}{“... not sufficient to establish conclusively}\\  \textcolor{teal}{that the person who possessed it so} \\ \textcolor{teal}{shortly after the murder of a person with} \\ \textcolor{teal}{whom he had enmity, had committed the} \\ \textcolor{teal}{murder...”}\end{tabular} \\ \cline{2-4} 
& BVA-Ext &                 \begin{tabular}[c]{@{}l@{}}.... under \textcolor{blue}{section I.C.}, by \\ the Session .... cattle shed in \\ village Bhadurpur Ghar. \textcolor{magenta}{With} \\ \textcolor{magenta}{respect to the learned Judges, } \\ \textcolor{magenta}{ theseobservations are not very} \\ \textcolor{magenta}{consistent. }\textcolor{teal}{To us, it seems that} \\ \textcolor{teal}{  in the middle of June when } \\ \textcolor{teal}{the chari and sugar cane crop } \\ \textcolor{teal}{would not have been very ......}\end{tabular} & \begin{tabular}[c]{@{}l@{}} \\ \textcolor{blue}{It mentions as I.C instead of I.P.C.}\\ \textcolor{blue}{and misses actual section number.} \\\  \\ \textcolor{magenta}{This sentence in the source document} \\ \textcolor{magenta}{is in context to Ujagar Singh’s} \\ \textcolor{magenta}{extrajudicial confession, and not the} \\ \textcolor{magenta}{details of the crime.}\\ \\ \textcolor{teal}{Incoherent discourse: It digresses} \\ \textcolor{teal}{into the details of the extrajudicial} \\ \textcolor{teal}{confession without describing the} \\ \textcolor{teal}{remaining evidence against the}\\  \textcolor{teal}{appellant}\end{tabular}                        \\ \hline
\end{tabular}
\caption{Examples of errors in summaries produced by BART model in zero-shot blind transfer (RQ1) for the Indian Supreme Court judgement \url{https://indiankanoon.org/doc/1811974/}, trained with different jurisdiction datasets.}
\label{error1}
\end{table*}

\begin{table*}[]
\centering
\begin{tabular}{|l|l|l|l|}
\hline
Config & Train & Summary Snippet & Explanation for Errors \\ \hline
& UK-Abs  & \begin{tabular}[c]{@{}l@{}} \\ ... \textcolor{blue}{under Section 302 K.P.C.} \\  ........................ murder. \\ \textcolor{magenta}{There is no good reasons for} \\ \textcolor{magenta}{Ujagar Singh to state falsely.} \\ If the statement made \\\\  \end{tabular} & \begin{tabular}[c]{@{}l@{}} \\ \textcolor{blue}{It mentions as K.P.C instead of I.P.C} \\ \\ \textcolor{magenta}{Incomplete sentence: It doesn’t mention} \\ \textcolor{magenta}{what Ujagar Singh states falsely.}\end{tabular}                                              \\ \cline{2-4} 
\multirow{-2}{*}{\begin{tabular}[c]{@{}c@{}}GRL \\ (RQ2)\end{tabular}}               & BVA-Ext & \begin{tabular}[c]{@{}l@{}} \\ \textcolor{blue}{section I.C.}, by the Session \\ Judge.................. the \\ appellant caused injuries to \\ Sheo Sahai, who was \\ sleeping in his cattle shed in \\ village Bhadurpur Ghar. \textcolor{teal}{With} \\ \textcolor{teal}{respect to the learned Judges,}\\ \textcolor{teal}{these observations are not very} \\ \textcolor{teal}{consistent.} The learned Sessions  \\ Judge has discussed the criticism \\ urged ...............it seems\\ that in the middle of June when \\ the chari and sugar cane crop \\ would not have been very \\ high, \textcolor{magenta}{it is improbable  Appeal} \\ \textcolor{magenta}{dismissed. }The evidence ........\\ \\ \end{tabular} & \begin{tabular}[c]{@{}l@{}}\textcolor{blue}{It mentions as I.C instead of I.P.C.}\\ \textcolor{blue}{and misses actual section number.} \\ \\ \textcolor{teal}{This sentence in the source document} \\ \textcolor{teal}{is in context to Ujagar Singh’s} \\ \textcolor{teal}{extrajudicial confession, and not the} \\  \textcolor{teal}{details of the crime.} \\ \\ \textcolor{magenta}{Incomplete sentence: It doesn’t} \\ \textcolor{magenta}{describe what is improbable and}\\ \textcolor{magenta}{moves to next sentence.} \end{tabular}  \\ \hline
& UK-Abs  & \begin{tabular}[c]{@{}l@{}} \\ Criminal Appeal No.\\ 1782 of 60 and Referred \\ No.125 of 1960, section 302 \\ I.P.C..........High Court erred \\ in rejecting the statement \\ of Ujagar Singh about the \\ appellant 's confessing \textcolor{magenta}{to him}\\ \textcolor{magenta}{that The extra judicial ....} \\ \\ \end{tabular} & \begin{tabular}[c]{@{}l@{}} \textcolor{magenta}{Incomplete sentence, which doesn’t} \\ \textcolor{magenta}{describe what the appellant confessed.}\end{tabular}   \\ \cline{2-4} 
\multirow{-2}{*}{\begin{tabular}[c]{@{}c@{}}Silver \\ (RQ3)\end{tabular}}            & BVA-Ext & \begin{tabular}[c]{@{}l@{}}\\ ... in \textcolor{blue}{Criminal Appeal} \\ \textcolor{blue}{No.125 of 1960,  section 302}\\  I.P.C........ for the appellant\\  has argued .......  \\ \\ \end{tabular}            & \begin{tabular}[c]{@{}l@{}} \\ \textcolor{blue}{Missing complete details: The source} \\ \textcolor{blue}{document states “...Criminal Appeal} \\ \textcolor{blue}{No. 1782 of 60 and Referred No. 125} \\ \textcolor{blue}{of 1960…” } \end{tabular}   \\ \hline
\end{tabular}
\caption{Examples of errors in summaries produced by BART model in GRL setting (RQ2) and silver summary setting (RQ3) for the Indian Supreme Court judgement \url{https://indiankanoon.org/doc/1811974/}, trained with different jurisdiction datasets.}
\label{error2}
\end{table*}